\documentclass[10pt,twocolumn,letterpaper]{article}

\usepackage{wacv}              

\definecolor{wacvblue}{rgb}{0.21,0.49,0.74}
\usepackage[pagebackref,breaklinks,colorlinks,allcolors=wacvblue]{hyperref}
\usepackage{siunitx}
\usepackage{xspace}
\usepackage{multirow}
\usepackage{xcolor}
\usepackage{colortbl}
\usepackage[T1]{fontenc} 
\usepackage{subcaption}
\usepackage{graphicx}
\usepackage{pifont}
\usepackage{enumitem}
\usepackage{geometry} 
\usepackage{hyperref}
\usepackage{lineno}
\usepackage[accsupp]{axessibility}  

\usepackage[protrusion=true,expansion=true]{microtype}
\def\wacvPaperID{1863} 
\def\confName{WACV}
\def\confYear{2026}

\newcommand{\datasetname}{ORCA\xspace}
\newcommand{\methodsCount}{18\xspace}
\newcommand{\tablealpha}{20}
\definecolor{class-level-color}{RGB}{4, 150, 255}
\definecolor{inter-class-color}{RGB}{255, 188, 66}
\definecolor{intra-class-color}{RGB}{216, 17, 89}
\definecolor{general-dataset-color}{RGB}{255,127,0}
\definecolor{marine-dataset-color}{RGB}{0, 79, 110}
\definecolor{open-set-color}{RGB}{106, 140, 175}
\definecolor{close-set-color}{RGB}{216, 184, 144}
\definecolor{image-level-color}{RGB}{122, 139, 82}
\definecolor{region-level-color}{RGB}{181, 115, 78}
\definecolor{zero-shot-color}{RGB}{75, 94, 148}
\definecolor{fine-tuned-color}{RGB}{210, 167, 104}

\title{ORCA: \underline{O}bject \underline{R}ecognition and \underline{C}omprehension for \\\underline{A}rchiving Marine Species}

\author{
Yuk-Kwan Wong$^{1}$ \quad
Haixin Liang $^{1}$ \quad
Zeyu Ma$^{2}$ \quad
Yiwei Chen$^{1}$ \quad
Ziqiang Zheng$^{1}$\thanks{Corresponding author: \texttt{zhengziqiang1@gmail.com}}\\
Rinaldi Gotama$^{3}$ \quad
Pascal Sebastian$^{3}$ \quad
Lauren D. Sparks$^{3}$ \quad
Sai-Kit Yeung$^{1}$\\[5pt]
$^{1}$Hong Kong University of Science and Technology\\
$^{2}$University of Electronic Science and Technology of China\quad
$^{3}$Indo Ocean Foundation\\[5pt]
\small Project website: \url{https://orca.hkustvgd.com}
}

\begin{document}

\let\oldtwocolumn\twocolumn
\renewcommand\twocolumn[1][]{%
    \oldtwocolumn[{#1}{
    \begin{center}
   \vspace{-0.3in}
   \includegraphics[width=\linewidth]{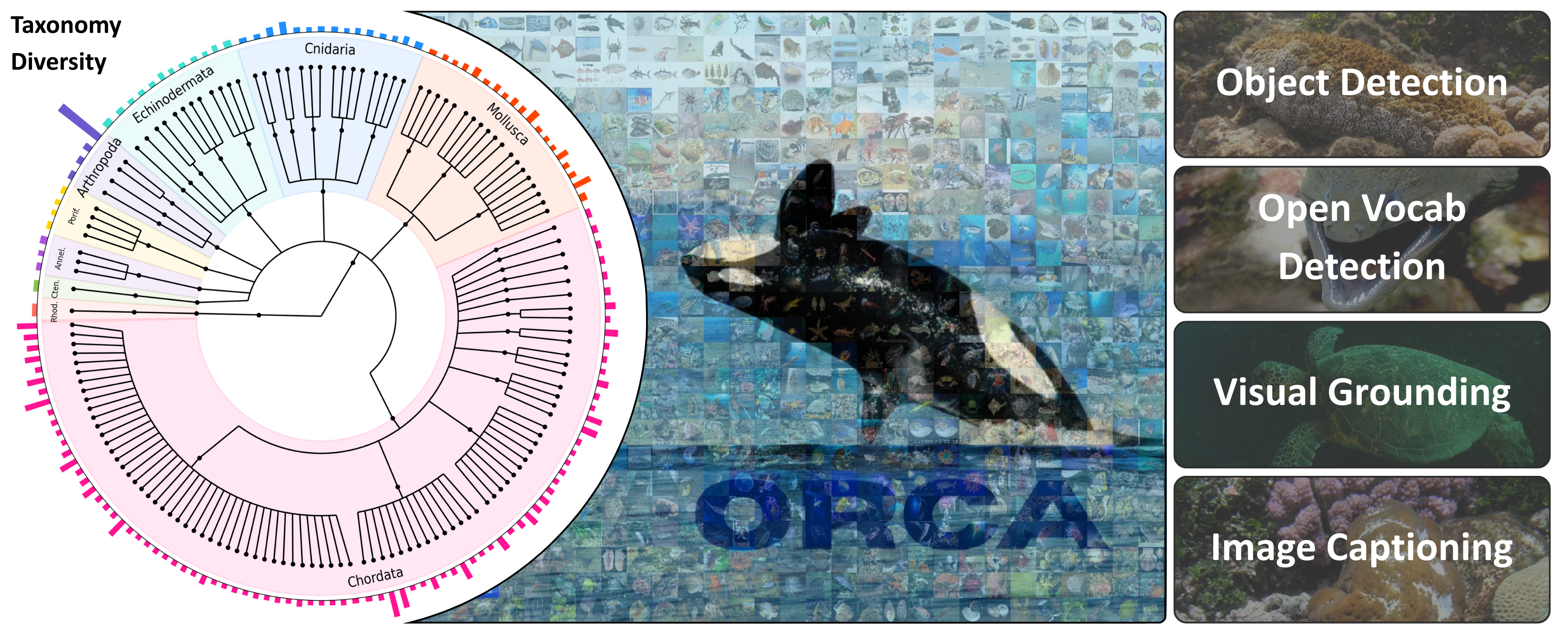}
   \captionof{figure}{\datasetname includes a spectrum of taxonomic marine species regarding both diversity and coverage. The detailed instance-level annotations (BBOX and dense captions enriched with domain-specific knowledge) enable object recognition and further comprehension for archiving the marine species, also supporting various biological applications.}
   \label{fig:teaser}
        \end{center}
    }]
}

\maketitle

\begin{abstract}
Marine visual understanding is essential for monitoring and protecting marine ecosystems, enabling automatic and scalable biological surveys. However, progress is hindered by limited training data and the lack of a systematic task formulation that aligns domain-specific marine challenges with well-defined computer vision tasks, thereby limiting effective model application. To address this gap, we present \datasetname, a multi-modal benchmark for marine research comprising 14,647 images from 478 species, with 42,217 bounding box annotations and 22,321 expert-verified instance captions. The dataset provides fine-grained visual and textual annotations that capture morphology-oriented attributes across diverse marine species. To catalyze methodological advances, we evaluate \methodsCount state-of-the-art models on three tasks: object detection (closed-set and open-vocabulary), instance captioning, and visual grounding. Results highlight key challenges, including species diversity, morphological overlap, and specialized domain demands, underscoring the difficulty of marine understanding. \datasetname thus establishes a comprehensive benchmark to advance research in marine domain.

\end{abstract}    
\section{Introduction}

The ocean, with a vast coverage on the surface of our blue planet, remains a mysterious abyssal region to the best of our knowledge. Advancing knowledge of marine ecosystems is critical for oceanography~\cite{Dong2022, Zhao2024}, sustainable resource management~\cite{10.1093/icesjms/fsab100, OCONNOR2024106133}, and biodiversity conservation~\cite{LOTZE2021R1190, Lionetto2021}. Considerable efforts have been devoted to biological surveys and habitat monitoring~\cite{zhang2024applications, yang2022uav, bae2023survey,wong2025coralscop}. To enhance the scalability and efficiency of in-situ monitoring, researchers are increasingly leveraging computer vision techniques to reduce manual data processing, including image classification~\cite{zhang2024catnet, stevens2024bioclip, demir2024projector}, object detection~\cite{zhang2021survey, yu2023multiple}, and vision–language modeling~\cite{li2025underwater}.

Despite the remarkable success enabled by powerful network backbones and domain-specific datasets, significant challenges remain, which can be broadly categorized into issues of \textbf{training data} and \textbf{task formulation}. Current marine datasets are often restricted to a small set of predefined categories (\emph{e.g.}, seven semantic classes in the UIIS dataset~\cite{Lian_2023_ICCV}) and are typically collected from limited geographic regions. Such constraints hinder both taxonomic diversity and ecological coverage, limiting the recognition of a broad range of marine species. Even fine-grained datasets with larger category sets, such as ~\cite{zhuang2018wildfish, zhuang2020wildfish++}, remain primarily focused on fish monitoring.

Regarding task formulation, current task definitions remain insufficient for domain requirements. \textit{Image-level classification}~\cite{zhang2024catnet, stevens2024bioclip, demir2024projector} may lead to inconsistency between coarse category annotations and image content, which often contains multiple species in a wild environment. \textit{Object detection}~\cite{zhang2021survey, yu2023multiple} is limited by a narrow set of categories. Furthermore, category labels alone cannot capture key biological traits that are essential for ecological monitoring. For \textit{image captioning}, although recent vision–language models (VLMs)~\cite{xu2024demystifyingclipdata,li2022blip,li2023blip2,zheng2023marinegpt} are evolving rapidly, their outputs are typically coarse and lack the granularity and domain-specific knowledge needed for description.

To address these challenges, we introduce \datasetname, the first multimodel dataset explicitly designed for marine research. \datasetname offers 1) \textbf{broad taxonomic coverage} spanning 478 species and 670 common-name categories; 2) \textbf{instance-level annotations} enabling both object detection and grounding; and 3) \textbf{biology-oriented captions} with diagnostic traits, appearances, behaviors, and habitats, all validated by marine biologists. The dataset comprises 14,647 images with 42,217 bounding boxes, each labeled with both scientific and common names to support diverse usage scenarios. In total, \datasetname provides 22,321 expertly verified instance–caption pairs, ensuring terminological accuracy and scientific relevance.

\datasetname supports a range of vision-language tasks, including closed-set and open-vocabulary detection, instance-level captioning, and visual grounding. While detection and grounding primarily assess a model’s ability to recognize and localize marine species, \datasetname further introduces three evaluation settings: \textit{Class-Level}, \textit{Intra-Class}, and \textit{Inter-Class}, to systematically examine how taxonomic hierarchies influence those abilities under the condition of \textbf{morphological overlapping}, where closely related species exhibit highly similar traits, thereby complicating species identification. Beyond spatial localization, the captioning and grounding components of \datasetname facilitate fine-grained alignment between visual observations and linguistic descriptions. This dual emphasis not only enhances object-level referencing but also supports the structured, biologically meaningful archiving of marine survey data. 

We have benchmarked \methodsCount state-of-the-art algorithms across the aforementioned tasks. In summary, our contributions can be outlined as follows:
\begin{itemize}[leftmargin=*, noitemsep, topsep=0pt]
    \item We present \datasetname, the first large-scale marine dataset with broad taxonomic coverage, bounding box annotations, and rich instance-level captions.
    \item We conduct an evaluation of \methodsCount models, showing that fine-tuning on \datasetname improves performance on localization and captioning tasks.
    \item  We demonstrate that dense, domain-specific captions enable accurate object referencing and resolve challenges posed by morphological overlap, where visual cues are ambiguous and misleading.
    \item We show that existing captioning models struggle with instance-level descriptions, often producing coarse, image-level captions instead of region-specific outputs.
\end{itemize}

\section{Related Work}

\noindent\textbf{Existing marine research}.  
Marine species exhibit high diversity in pose, appearance, and pattern. Robust marine visual understanding can leverage recent algorithms~\cite{li2021marine,hong2020trashcan,ziqiang2024coralscop,ziqiang2025coralsrt} to advance research, conservation, and industry. Several datasets have been introduced, including MAS3K~\cite{li2020mas3k,li2021marine}, WildFish~\cite{zhuang2018wildfish}, WildFish++~\cite{zhuang2020wildfish++}, and SUIM~\cite{islam2020semantic}, which improve recognition of marine organisms. However, most of them provide only a limited set of predefined categories without detailed captions, restricting their utility for fine-grained marine analysis and large-scale scientific databases. \datasetname addresses this gap by introducing a large-scale dataset covering a broad range of marine species with high-quality annotations (bounding boxes and captions).

\noindent\textbf{Object Detection}.  
Object detection is a core computer vision task~\cite{lin2014microsoft,ren2015faster,ren2016faster}, involving simultaneous object localization and classification. Conventional one-stage~\cite{liu2016ssd,ge2021yolox,redmon2016you} and two-stage~\cite{ren2015faster,ren2016faster,he2017mask} detectors rely on fixed predefined category sets, which limit their applicability in marine domains, where species diversity varies greatly across regions. Open-vocabulary object detection (OVOD)~\cite{zareian2021open,yao2023detclipv2,kim2023region,wang2023detecting} addresses this challenge by extending detection to unseen categories. OVOD commonly leverages large-scale vision–language pre-training~\cite{radford2021learning} to align visual regions with textual concepts; for instance, RegionCLIP~\cite{kim2023region} enhances generalization by matching regional features with natural language. These properties make OVOD particularly promising in marine applications, with the ability to recognize novel and diverse species.

\noindent\textbf{Vision–Language Understanding}.  
VLMs~\cite{liu2024visual,2023GPT4VisionSC,team2023gemini,zhu2023minigpt,liu2023visual,zheng2023marinegpt,li2022blip,li2023blip,ziqiang2024marineinst} have made substantial progress, driven by large-scale datasets such as Visual Genome~\cite{krishna2017visual}, VizWiz~\cite{Gurari2018VizWizGC}, RefCOCO~\cite{kazemzadeh2014referitgame}, and Objects365~\cite{shao2019objects365}. These models combine visual encoders~\cite{dosovitskiy2020image} with large language models~\cite{chatgpt,openai2023gpt4}, trained on massive image–text corpora. CLIP~\cite{radford2021learning} demonstrated strong zero-shot recognition, while BLIP~\cite{li2022blip,li2023blip} advanced multimodal pre-training through frozen encoder–decoder architectures. Collectively, these works provide the foundation for tasks such as image captioning and grounding, which are critical for automatically documenting and archiving marine observations and discoveries. However, most existing datasets focus on terrestrial objects with very limited marine coverage, restricting VLM effectiveness in this domain. Furthermore, current VLMs struggle with fine-grained, region-level instance understanding essential for marine-specific tasks. To address this gap, \datasetname\ provides high-quality textual annotations to better enable VLM applications in marine research.

\section{Orca Construction}
\label{sec:method}

\begin{figure*}[h]
    \centering
    \includegraphics[width=\linewidth]{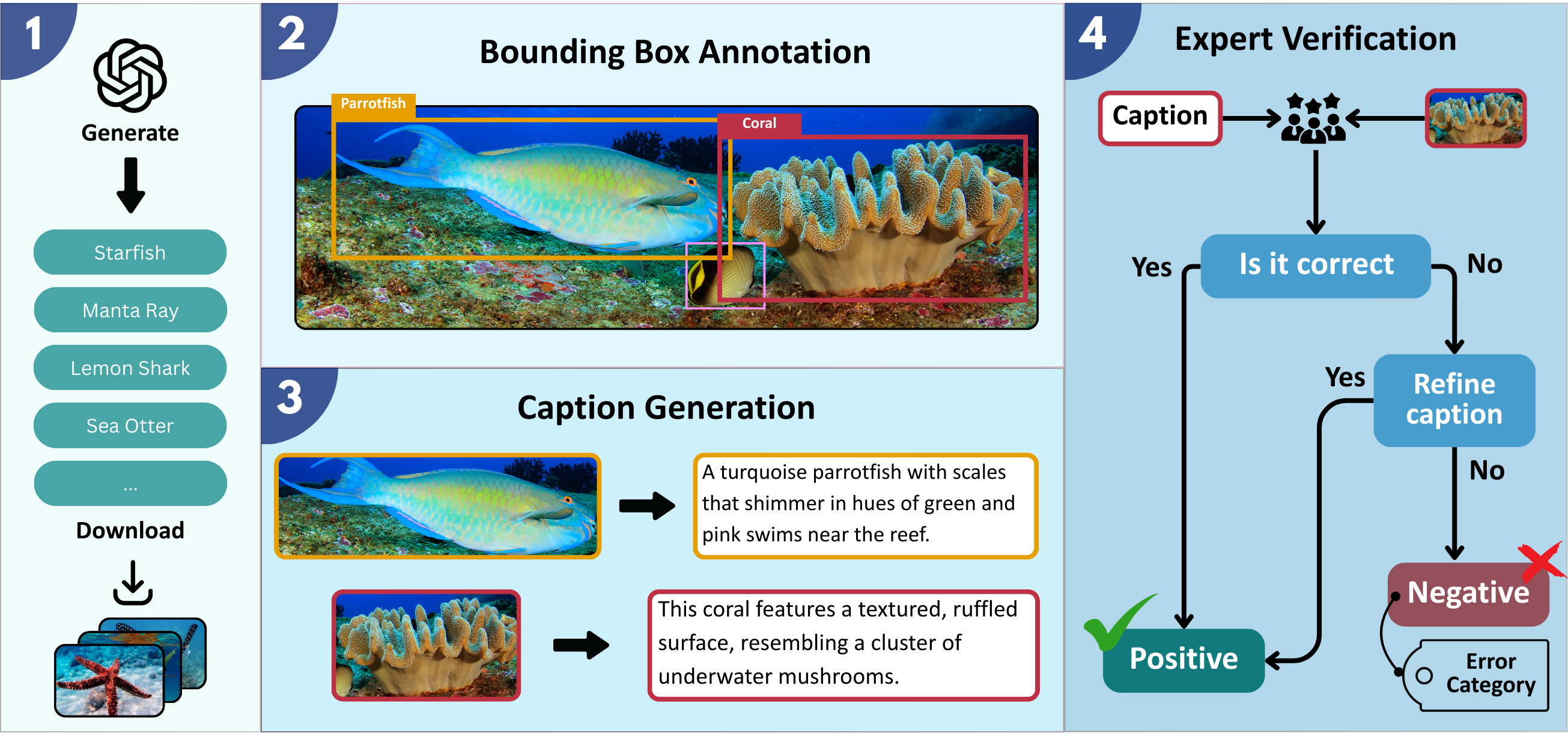}
    \caption{Overview of \datasetname construction process. It begins with image collection, followed by bounding box annotation and caption generation for each box. Domain experts then verify all of them and refine at least one caption per image.}
    \label{fig:enter-label}
    \vspace{-0.1in}
\end{figure*}

We illustrate the construction protocol of \datasetname in Figure~\ref{fig:enter-label} and subsequently summarize its characteristics and statistics.

\subsection{Dataset Construction}
\label{sec:dataset_construction}

\textbf{Data collection}. 
The process began by compiling a target list of marine taxonomic categories. GPT-4 was employed to generate canonical common names (\emph{e.g.}, seahorse), providing a proxy for vernacular terms most widely used by the public and thereby guiding more effective image searches. Candidate images were then sourced from Google Images, Flickr, and iNaturalist, with URLs retained for copyright attribution. All images underwent manual inspection to remove duplicates and misclassified entries, ensuring both quality and diversity. Each common name was subsequently mapped to its corresponding taxon in the World Register of Marine Species (WoRMS)~\cite{WoRMS}. Cases where a common name referred to an entire genus or higher taxonomic rank (\emph{e.g.}, ``unicorn fish,'' encompassing the genus \textit{Naso}) were excluded to avoid ambiguity.

\begin{figure}[t]
    \centering
    \includegraphics[width=\linewidth]{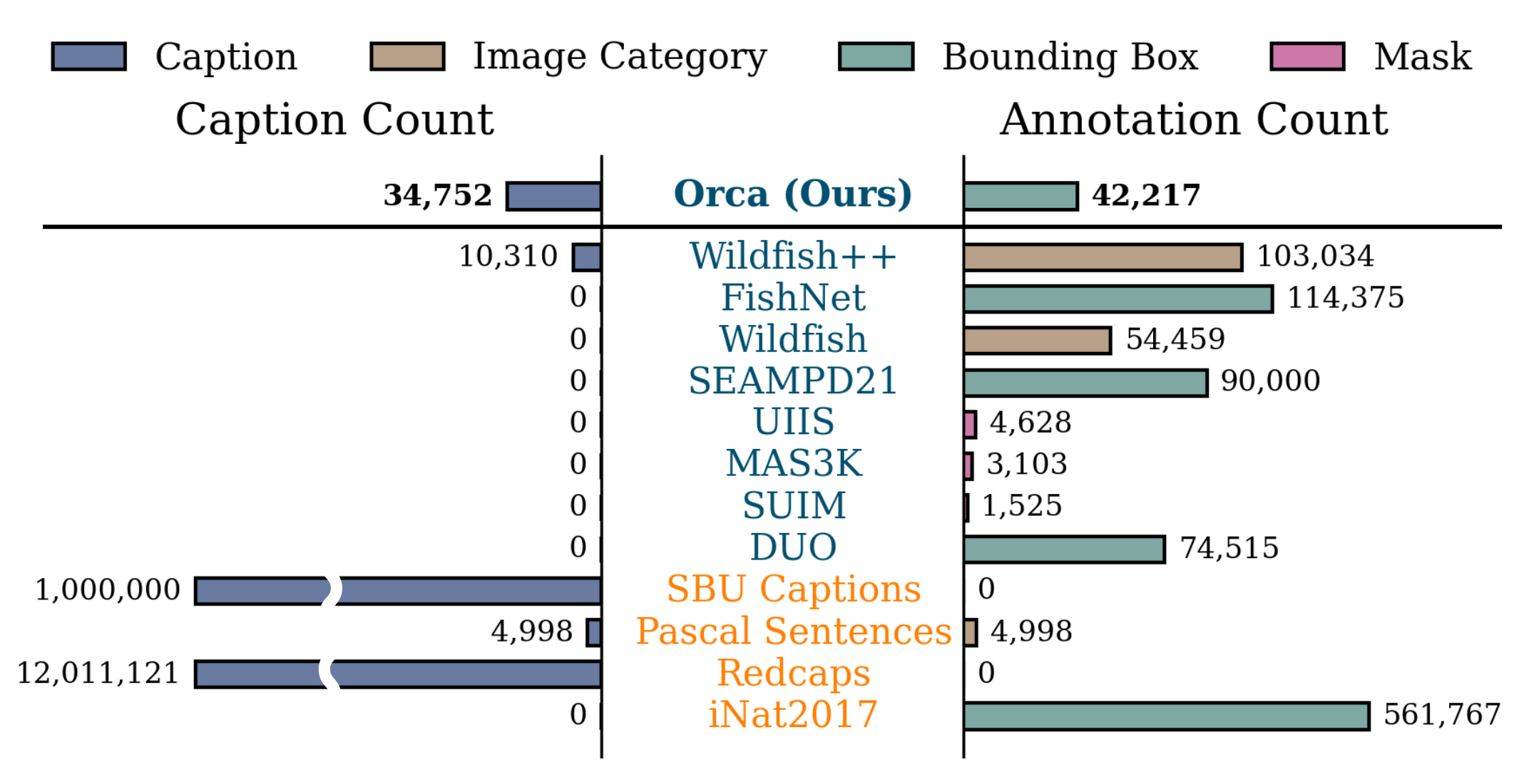}
    \caption{\datasetname offers a balanced and sufficient amount of visual and textual annotations, compared to \textcolor{general-dataset-color}{general} and \textcolor{marine-dataset-color}{domain-specific} datasets.}\label{fig:annotation_quantities}
\end{figure}

\noindent\textbf{Bounding-box annotation.} 
We combined the Segment Anything Model (SAM) \cite{kirillov2023segment} with human-supplied point prompts to delineate object masks, which were subsequently converted to axis-aligned bounding boxes. Given the amorphous morphology of marine organisms, we specifically verified that each box fully encompassed the target instance, including translucent fins and slender appendages.

\noindent\textbf{Caption generation}. Existing datasets~\cite{schuhmann2021laion} mainly utilized alt-texts to formulate the image-text pairs. However, the texts suffer from limited information (short captions), misalignment with visual contents, and deviation from domain-specific requirements. Instead, we generate rich instance-level descriptions. For every large bounding box ($>$ \SI{1024}{pixels}), the cropped region is passed to MarineGPT~\cite{zheng2023marinegpt} to produce captions tailored to the marine research.

\begin{table}[t]
\centering
\scalebox{0.5}{
\begin{tabular}{l|c|cc|cc}
\hline
\multicolumn{1}{c|}{Dataset} & \begin{tabular}[c]{@{}c@{}}Image \\ Count\end{tabular} & \begin{tabular}[c]{@{}c@{}}Visual\\ Annotation\end{tabular} & \begin{tabular}[c]{@{}c@{}}Lingistic\\ Annotation\end{tabular} & \begin{tabular}[c]{@{}c@{}}Category\\ Count\end{tabular} & \begin{tabular}[c]{@{}c@{}}Taxonomy\\ Supported\end{tabular} \\ \hline
\textcolor{marine-dataset-color}{DUO}~\cite{liu2021dataset} & 7,782 & BBOX & - & 4 & \ding{55} \\
\textcolor{marine-dataset-color}{SUIM}~\cite{islam2020semantic} & 1,525 & Mask & - & 8 & \ding{55} \\
\textcolor{marine-dataset-color}{MAS3K}~\cite{li2020mas3k} & 3,103 & Mask & - & 37 & \ding{55} \\
\textcolor{marine-dataset-color}{UIIS}~\cite{Lian_2023_ICCV} & 4,628 & Mask & - & 7 & \ding{55} \\
\textcolor{marine-dataset-color}{SEAMPD21}~\cite{boulais2021seamapd21} & 28,328 & BBOX & - & 130 & \ding{55} \\
\textcolor{marine-dataset-color}{Wildfish}~\cite{zhuang2018wildfish} & 54,459 & Category & - & 1,000 & \ding{55} \\
\textcolor{marine-dataset-color}{FishNet}~\cite{Khan_2023_ICCV} & 94,532 & BBOX & - & 17,357 & \ding{51} \\
\textcolor{marine-dataset-color}{Wildfish++}~\cite{zhuang2020wildfish++} & 2,348 & Category & Image-Level & 2,348 & \ding{51} \\ \hline
\textcolor{general-dataset-color}{Redcaps}~\cite{desai2021redcaps} & 12,011,121 & - & Image-Level & - & \ding{55} \\
\textcolor{general-dataset-color}{Pascal Sentences}~\cite{rashtchian-etal-2010-collecting} & 1,000 & Category & Image-Level & 20 & \ding{55} \\
\textcolor{general-dataset-color}{SBU Captions}~\cite{Ordonez:2011:im2text} & 1,000,000 & - & Image-Level & - & \ding{55} \\ 
\textcolor{general-dataset-color}{iNat2017~\cite{vanhorn2018inaturalistspeciesclassificationdetection}} & 859,000 & BBOX & - & 5,089 & \ding{51} \\
\hline
\textcolor{marine-dataset-color}{Orca (Ours)} & 14,645 & BBOX & \textbf{Instance-Level} & 670 & \ding{51} \\ \hline
\end{tabular}
}

\caption{Statistic comparison with other \textcolor{general-dataset-color}{general} and \textcolor{marine-dataset-color}{domain-specific} datasets.}
\label{tab:dataset_statistic}
\end{table}

\vspace{-0.05in}
\subsection{Dataset Statistic and Comparison}
\label{sec:dataset_statistic}

\noindent\textbf{Caption refinement}. The generated caption is then passed to domain experts for verification and refinement along four dimensions: 1) \textit{Unique morphological traits}, such as color, shape, injuries, \textit{etc}; 2) \textit{Spatial context} (absolute and relative positions); 3) \textit{Environmental background}; and 4) \textit{Behavioral cues} (individual or inter-species interactions). To enhance labeling efficiency, experts are required to refine at least one caption per image. The remaining are labeled \textit{positive} if error-free or \textit{negative} otherwise. We intentionally retained the \textit{negative} captions, proving harder negatives than prior work that substitutes random nouns~\cite{zhao2022vl, yuksekgonul2022and}. Finally, \datasetname contains \SI{34752}{} captions (with \SI{12873}{} \textit{refined}, \SI{9448}{} \textit{positive} and \SI{12431}{} \textit{negative} captions). We further codify 11 error categories responsible for negative labels, where details are provided in the supplementary material.

\begin{figure}[t]
    \centering
    \includegraphics[width=0.95\linewidth]{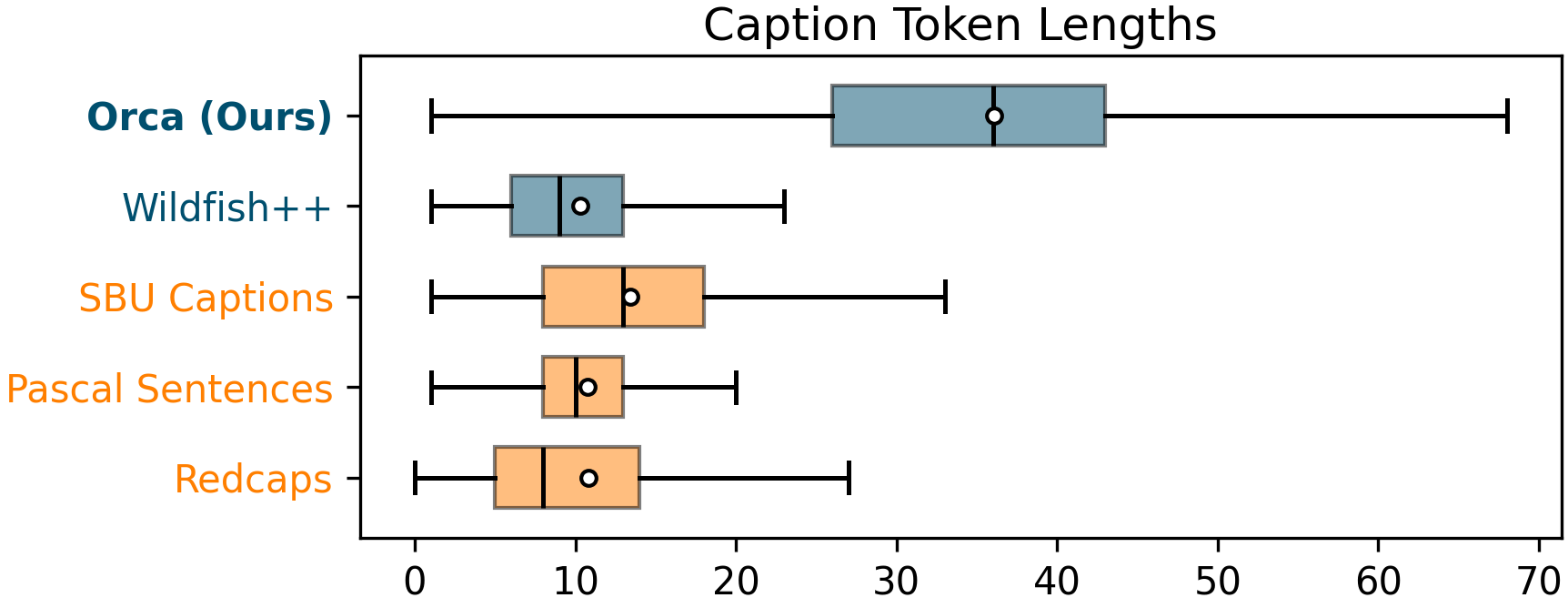}
    \caption{Caption tokens length for \textcolor{general-dataset-color}{general datasets} and \textcolor{marine-dataset-color}{domain-specific datasets}. The white circle represents the mean caption length. Outliers have been filtered out.}
    \label{fig:caption_length}
\end{figure}

\begin{figure}[t]
    \centering
    \small 
    \begin{subfigure}{0.29\linewidth} 
        \centering
        \includegraphics[width=\linewidth]{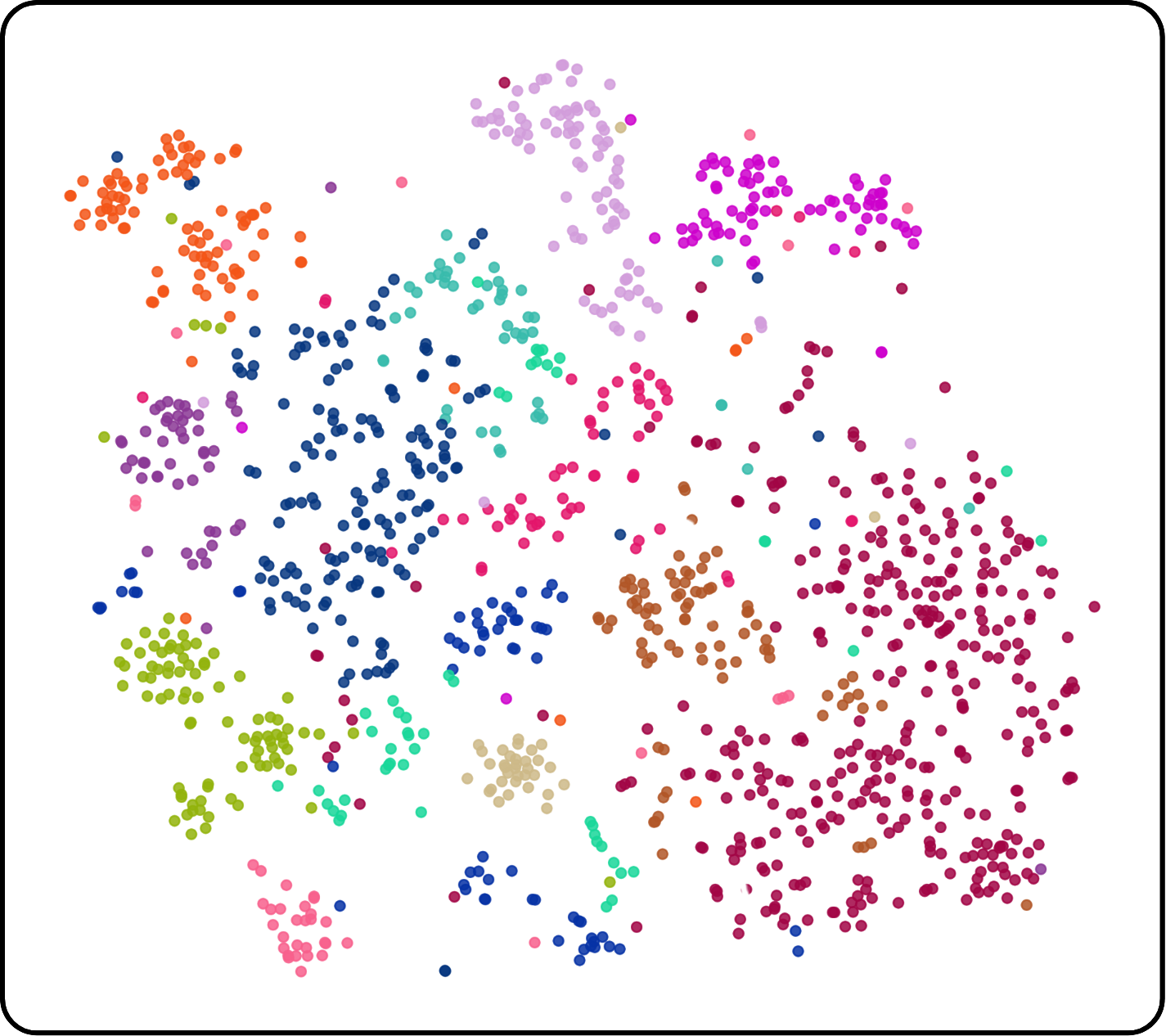}
        \caption{\textcolor{general-dataset-color}{Pascal Sent.}}
    \end{subfigure}
    \hspace{0.02\linewidth} 
    \begin{subfigure}{0.29\linewidth} 
        \centering
        \includegraphics[width=\linewidth]{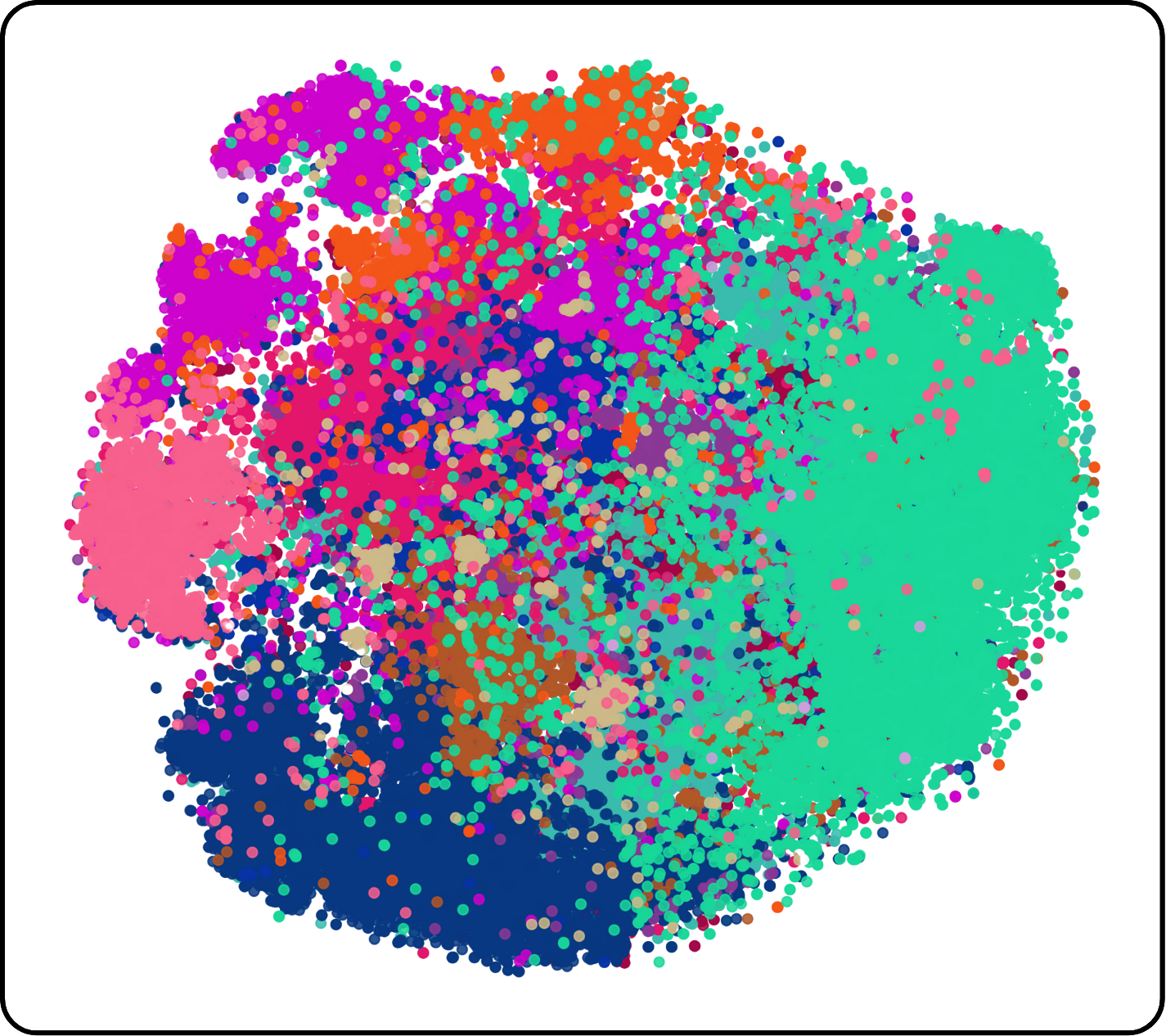}
        \caption{\textcolor{general-dataset-color}{Redcaps}}
    \end{subfigure}
    \hspace{0.02\linewidth} 
    \begin{subfigure}{0.29\linewidth} 
        \centering
        \includegraphics[width=\linewidth]{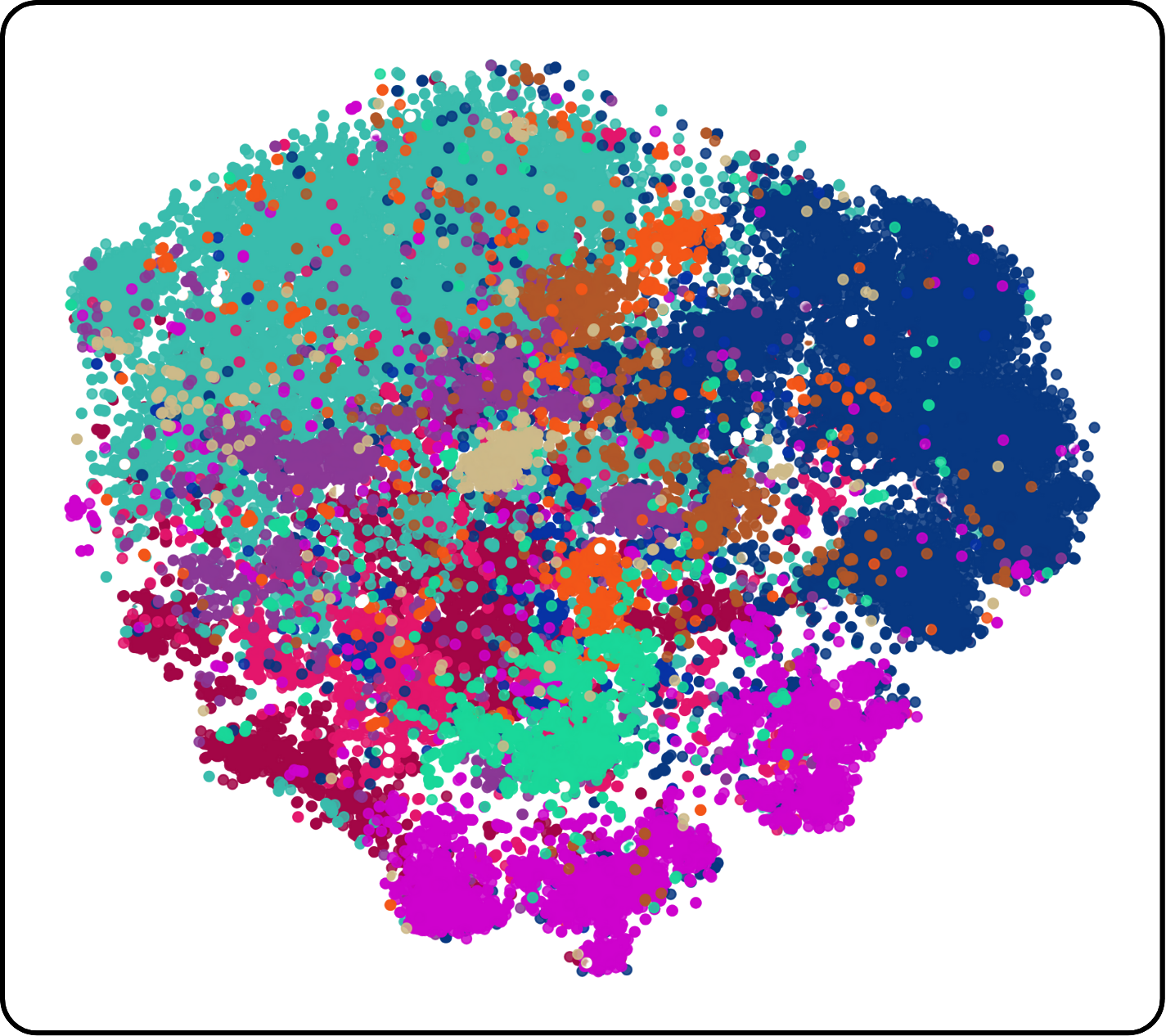}
        \caption{\textcolor{general-dataset-color}{SBU Captions}}
    \end{subfigure}

    \begin{subfigure}{0.29\linewidth} 
        \centering
        \includegraphics[width=\linewidth]{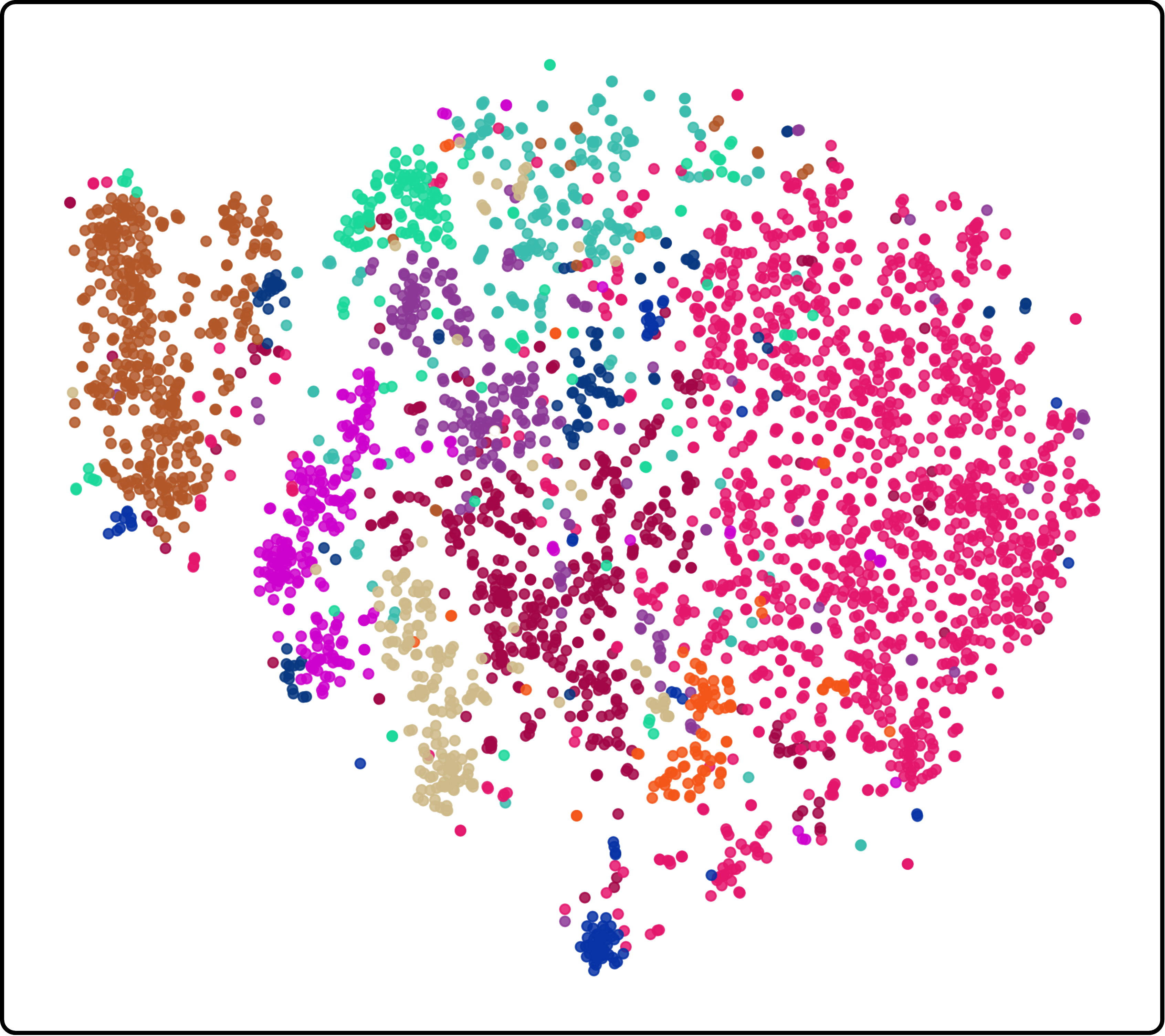}
        \caption{\textcolor{marine-dataset-color}{\datasetname (Ours)}}
    \end{subfigure}
    \hspace{0.1\linewidth} 
    \begin{subfigure}{0.29\linewidth} 
        \centering
        \includegraphics[width=\linewidth]{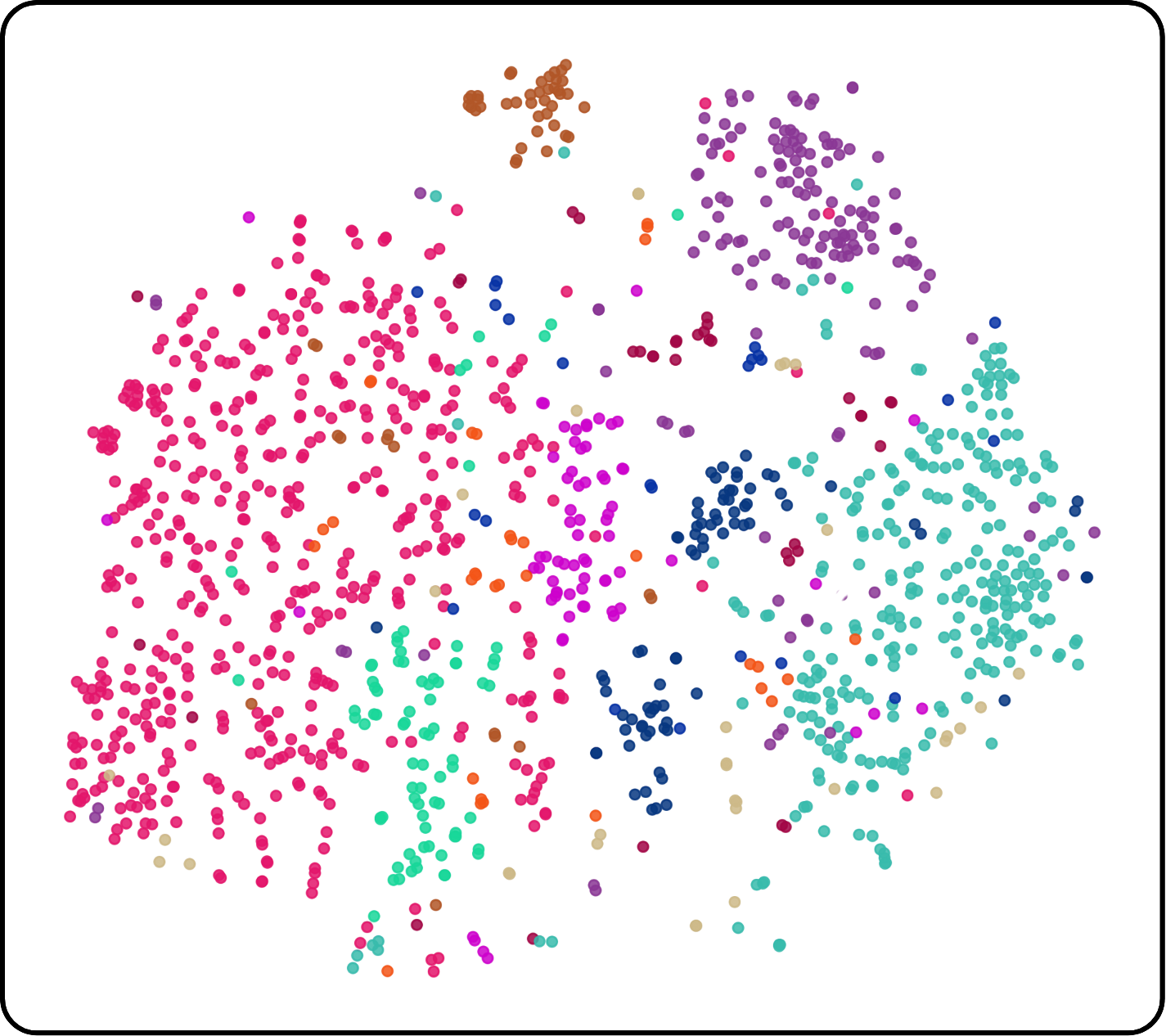}
        \caption{\textcolor{marine-dataset-color}{Wildfish++}}
    \end{subfigure}
    \caption{t-SNE of the vocabulary used in \textcolor{general-dataset-color}{general datasets} and \textcolor{marine-dataset-color}{domain-specific datasets}). \textcolor{general-dataset-color}{Pascal Sent.} stand for \textcolor{general-dataset-color}{Pascal Sentence} for better visualization.}
    \label{fig:word_diversity}

\end{figure}

Our dataset introduces domain-specific features that distinguish it from both general-purpose and existing marine datasets:  1) it provides comprehensive instance-level annotations, with each bounding box larger than \SI{1024}{pixels} paired with a caption and mapped to marine taxonomic categories, as shown in Table~\ref{tab:dataset_statistic}; 2) it ensures balanced visual–textual supervision, offering comparable scales across both modalities to support a wide range of vision–language tasks, unlike other datasets that emphasis on one modality, as illustrated in Figure~\ref{fig:annotation_quantities}; and 3) it includes dense and diverse captions for each organism, yielding high caption density and substantial vocabulary diversity, as shown in Figure~\ref{fig:caption_length} and Figure~\ref{fig:word_diversity} respectively.

\section{Experiments}

We benchmark \methodsCount existing SOTA models on \datasetname from three representative tasks, including object detection (closed-set and open-vocabulary settings), instance captioning, and visual grounding.

\subsection{Object Detection}
\label{sec:object_detection}

\noindent\textbf{Experimental settings}. 
We evaluate the capacity of closed-set and open-vocabulary object detection models to both localize and identify marine creatures. It is notoriously challenging, even for experienced biologists, because of the morphological overlap among species, where those belonging to the same higher-level taxon often exhibit similar physical characteristics. For OVOD, we devise three settings: \textit{Class-Level}, \textit{Intra-Class}, and \textit{Inter-Class}.

\noindent\textbf{Class-Level}. We group species at ``Class'' level in the taxonomic hierarchy. Specifically, 670 vernacular categories are consolidated into 33 Class-level taxonomic categories, with 24 \textit{seen} and 9 \textit{unseen} categories. Certain vernacular categories (\emph{e.g.}, \textit{Bryozoa}) correspond to higher taxonomic ranks (\emph{e.g.}, \textit{phylum}) and are therefore excluded. 

\noindent\textbf{Intra-Class} setting refines the task further by requiring models to identify vernacular categories within the aforementioned 33 Class-level taxonomic groups. From these, we sample 555 vernacular categories as \textit{seen} and 109 as \textit{unseen}.

\noindent\textbf{Inter-Class}. We adopt a more granular approach by sampling one vernacular category as \textit{unseen} for every four categories within each ``Class'', while designating the remaining three vernacular categories as \textit{seen}. ``Classes'' with fewer than four categories are excluded. As a result, this setup includes 482 \textit{seen} and 161 \textit{unseen} vernacular categories.

\begin{table*}[t]
  \centering
  \footnotesize
   \scalebox{1.0}{
   \begin{tabular}{l|ccc|ccc}
    \toprule
    \multirow{2}{*}{Method}  & \multicolumn{3}{c|}{Seen}  & \multicolumn{3}{c}{Unseen}  \\
    & Class-level & Intra-Class & Inter-Class & Class-level & Intra-Class & Inter-Class \\
        \midrule
    \rowcolor{close-set-color!\tablealpha} FasterRCNN~\cite{ren2015faster}  & 28.7 & 17.6 & 16.7 & - & - & -    \\
    \rowcolor{close-set-color!\tablealpha} YOLOX~\cite{ge2021yolox} & 27.5 &  21.7  & 21.0  & - & - & -    \\
    \rowcolor{close-set-color!\tablealpha} GridRCNN~\cite{lu2019grid} & 32.7 & 28.1  &  28.6 & - & - & -    \\
    \midrule
    \rowcolor{open-set-color!\tablealpha} UniDetector~\cite{wang2023detecting} & 31.5 & 23.3 & 24.1 & 8.2 & 0.4 &  \underline{0.7}   \\
    \rowcolor{open-set-color!\tablealpha} RegionCLIP~\cite{zhong2022regionclip} & \underline{39.8} & \underline{34.1} & \underline{29.8} & \underline{12.2} & \underline{6.2} & 0.4  \\
    \rowcolor{open-set-color!\tablealpha} DECOLA~\cite{cho2023language}& \textbf{66.7} & \textbf{88.8} & \textbf{86.9} & \textbf{37.7} & \textbf{51.6} &   \textbf{52.3}   \\
    \bottomrule
    \end{tabular}
   }
 
\caption{Quantitative object detection results (mAP$_{50}$) under \textcolor{close-set-color}{close-set} and \textcolor{open-set-color}{open-vocabulary} settings.}
 \label{tab:detection-table}
 \vspace{-0.2in}
\end{table*}

\noindent\textbf{Close-set object detection}. We mainly include 3 representative close-set object detection algorithms (Faster-RCNN~\cite{ren2015faster}, YOLOX~\cite{ge2021yolox}, and GridRCNN~\cite{lu2019grid}) and report the mAP$_{50}$ of 24 \textit{seen} categories under three settings. Our implementation of these models is based on MMDetection~\cite{chen2019mmdetection} using the official experimental setting. Please note that we do not evaluate these closed-set object detection algorithms on the \textit{unseen} categories. 

\noindent\textbf{OVOD}. We evaluate the performance of 3 open-vocabulary object detection algorithms (UniDetector~\cite{wang2023detecting}, RegionCLIP~\cite{zhong2022regionclip}, and DECOLA~\cite{cho2023language}). We follow the official experimental setting and fine-tune the model on our \datasetname dataset. Particularly, we adopt the single-dataset training strategy for UniDetector~\cite{wang2023detecting} to continuously optimize it in an end-to-end fashion. For DECOLA~\cite{cho2023language}, we utilize their best-performing model with Swin-B backbone (phase 1) as the pre-trained model. We inherit the language-conditioned detection training procedure of DECOLA while keeping other configurations the same. We report the quantitative result in Table~\ref{tab:detection-table} where mAP$_{50}$ is computed.

\noindent\textbf{Comparison and analysis.} 
Detecting marine organisms poses significant challenges for general object detection models. As summarized in Figure~\ref{fig:quantitative_results} and Table~\ref{tab:detection-table}, while these models effectively locate objects, they struggle with accurate classification. We summarize two observations: 1) Morphological overlap across species confuses models for species identification, resulting in lower performance for Intra- and Inter-Class compared to Class-Level. 2) Relying solely on visual cues is insufficient for species identification. Performance in closed-set object detection (which relies exclusively on visual features) is generally lower than in open-vocabulary object detection (which incorporates both visual features and category labels). DECOLA demonstrates a clear advantage in recognizing fine-grained marine species. We attribute this to its language-conditioned query selection strategy.

\subsection{Instance Captioning}
\label{sec:image_caption}

\begin{figure*}[t]
    \centering
    \includegraphics[width=\textwidth]{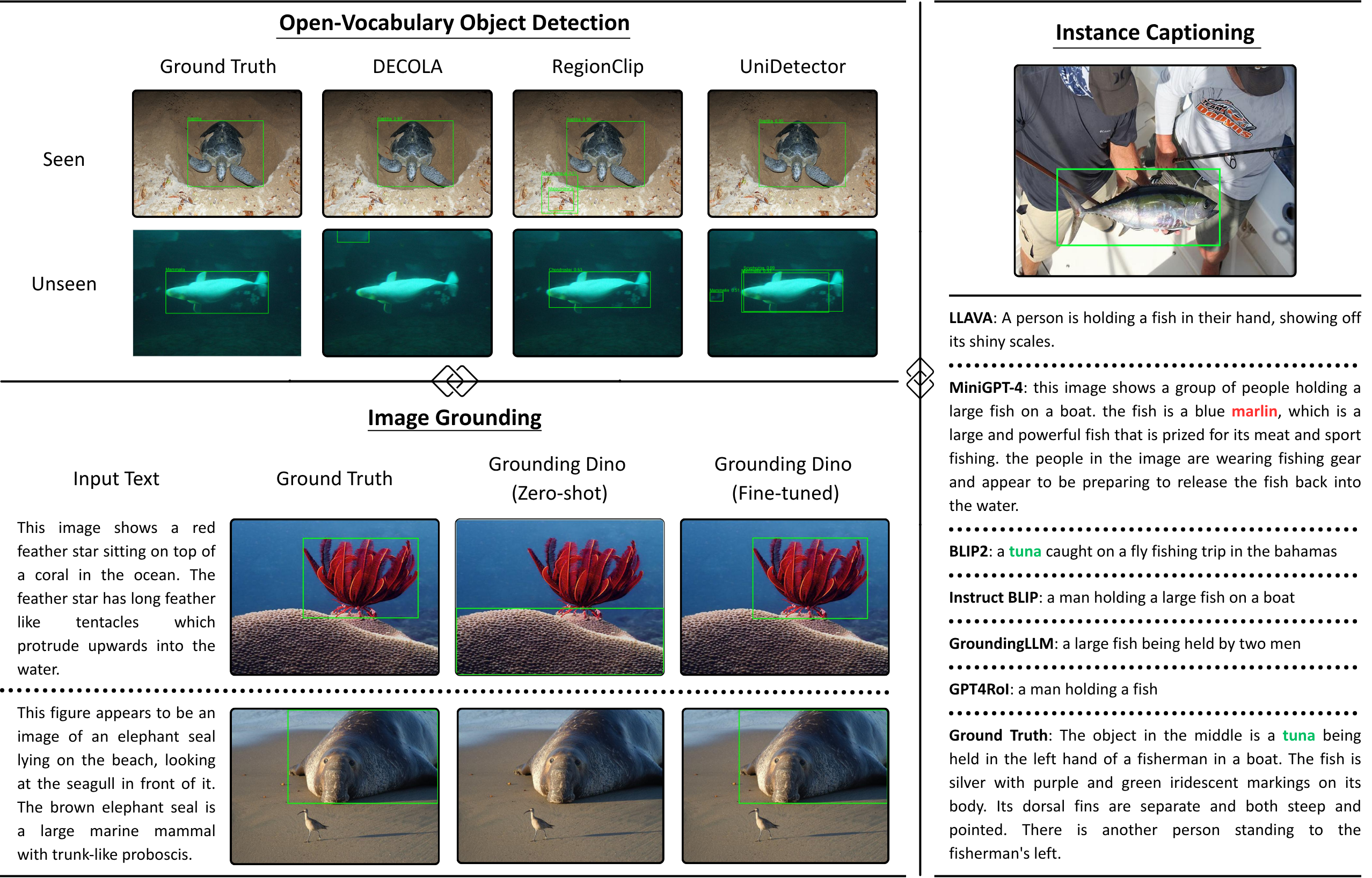}
    \caption{Quantitative results of open-vocabulary object detection, visual grounding, and image captioning.}
    \label{fig:quantitative_results}
 
\end{figure*}

\begin{table*}
  \centering
  \footnotesize
  \scalebox{1.0}{
  \begin{tabular}{l|cccccc}
    \toprule
    Method & CLIPScore$\uparrow$ & RefCLIPScore$\uparrow$ & CIDEr$\uparrow$ & BLUE-4$\uparrow$ & METEOR$\uparrow$ & ROUGE$\uparrow$ \\
        \midrule
    \rowcolor{image-level-color!\tablealpha} LLAVA~\cite{liu2024visual} & 73.78 & 72.27 & 4.93 & \underline{8.77} & \underline{7.70} & 20.76 \\
  
    \rowcolor{image-level-color!\tablealpha} MiniGPT-4~\cite{zhu2023minigpt} & 74.48 & 73.43 & 5.72 & 7.18 & \textbf{16.90} & \textbf{28.03} \\
    \rowcolor{image-level-color!\tablealpha} BLIP2~\cite{li2023blip2} & \underline{76.22} & \underline{73.73} & \underline{9.96} & 8.16 & 5.95 & 18.96 \\
    \rowcolor{image-level-color!\tablealpha} InstructBLIP~\cite{dai2024instructblip} & \textbf{76.60} & \textbf{75.25} & \textbf{12.09} & \textbf{13.94} & 7.40 & \underline{21.31} \\
    \midrule
    \rowcolor{region-level-color!\tablealpha} GroundingLMM {\scriptsize{(RefCOCOg)}}~\cite{rasheed2023glamm} & 73.04 & 70.97 & 4.37 & 4.39 & 4.60 & 16.37 \\
    \rowcolor{region-level-color!\tablealpha} GroundingLMM (VG)~\cite{rasheed2023glamm} & 71.15 & 69.04 & 4.06 & 2.47 & 4.11 & 15.22 \\
    \rowcolor{region-level-color!\tablealpha} GPT4RoI~\cite{zhang2023gpt4roi} & 71.28 & 68.71 & 3.53 & 2.81 & 4.07 & 15.08 \\
    \bottomrule
    \end{tabular}
  }

    \caption{Results of various algorithms (\textcolor{image-level-color}{image-level} and \textcolor{region-level-color}{region-level}) on instance captioning.} 
  \label{tab:instance_capations}
  \vspace{-0.2in}

\end{table*}

\noindent\textbf{Experimental settings}. We benchmark off-the-shelf VLMs from two aspects: image-level and region-level. The former image-level VLMs (LLAVA~\cite{liu2024visual}, MiniGPT-4~\cite{zhu2023minigpt}, BLIP2~\citep{li2023blip2}, and InstructBLIP~\citep{dai2024instructblip}) were optimized by image-level captions and lacked the ability to understand specific object instances. We evaluate these image-level VLMs based on the following user instruction: ``\textit{describe the object in this figure}''. The latter region-level VLMs (GroundingLMM~\cite{rasheed2023glamm}, GPT4RoI~\cite{zhang2023gpt4roi}) were optimized by paired image region prompts and the corresponding instance captions. We provide the BBOX annotation in the given text prompt following the experimental setting of ~\cite{rasheed2023glamm,zhang2023gpt4roi}. We perform the evaluations based on expert-verified instance captions to analyze their capability in describing marine instance objects. To quantitatively measure the performance of various algorithms, we adopt the widely used captioning metrics (including CLIPScore, RefCLIPScore~\cite{hessel2021clipscore}, CIDEr~\cite{vedantam2015cider}, BLUE-4~\cite{papineni2002bleu}, METEOR~\cite{banerjee2005meteor}, and Rouge~\cite{lin2004rouge}) to compute quantitative results in Table~\ref{tab:instance_capations}. Besides the human-constructed instance captions proposed in \datasetname, we also construct a starting sentence to include the category information for the selected object instance: ``This is a <\textit{Category Name}>.'', where the <\textit{Category Name}> is the placeholder to compensate for the scientific category-level information of each instance. In this way, by penalizing generated plausible but not domain-specific responses (\emph{e.g.,} ``fish'', ``animal'', and ``mammal''), we encourage the model to generate the scientific captions to satisfy the domain requirements.

\begin{table*}[t]
\centering
\footnotesize
\scalebox{0.9}{
\begin{tabular}{lcccccc }
\hline
\multicolumn{7}{c}{\textcolor{zero-shot-color}{\textit{Zero-shot}}} \\ \hline
\multicolumn{1}{l|}{} &
  \multicolumn{6}{c}{Unseen} \\
\multicolumn{1}{l|}{\multirow{-2}{*}{Method}} &
  Class-Level &
  Intra-Class &
  \multicolumn{1}{c|}{Inter-Class} &
  Class-Level &
  Intra-Class &
  Inter-Class \\ \hline
\rowcolor{zero-shot-color!\tablealpha} 
\rowcolor{zero-shot-color!\tablealpha} \multicolumn{1}{l|}{GroundingVLP~\cite{shen2023groundvlp}} &
  0.5183 &
  0.5148 &
  \multicolumn{1}{c|}{0.518} &
  0.5816 &
  0.5543 &
  0.5837 \\
\rowcolor{zero-shot-color!\tablealpha} \multicolumn{1}{l|}{TransVG~\cite{deng2021transvg}} &
  0.5191 &
  0.5025 &
  \multicolumn{1}{c|}{0.5048} &
  0.5849 &
  0.5492 &
  0.5773 \\
\rowcolor{zero-shot-color!\tablealpha} \multicolumn{1}{l|}{GroundingDino~\cite{liu2023grounding}} &
  0.5674 &
  0.5606 &
  \multicolumn{1}{c|}{0.5853} &
  0.6324 &
  0.5868 &
  0.6278 \\
\rowcolor{zero-shot-color!\tablealpha} \multicolumn{1}{l|}{HiVG~\cite{xiao2024hivg}} &
  0.4751 &
  0.4386 &
  \multicolumn{1}{c|}{0.4471} &
  0.5459 &
  0.4743 &
  0.5399 \\
\rowcolor{zero-shot-color!\tablealpha} \multicolumn{1}{l|}{Dynamic-MDETR~\cite{shi2024dynamic}} &
  0.5261 &
  0.5004 &
  \multicolumn{1}{c|}{0.5092} &
  0.5856 &
  0.5484 &
  0.5792 \\
\rowcolor{zero-shot-color!\tablealpha} \multicolumn{1}{l|}{CLIP-VG~\cite{xiao2023clip}} &
  0.5499 &
  0.5357 &
  \multicolumn{1}{c|}{0.5346} &
  0.6281 &
  0.5789 &
  0.6233 \\ \hline
\multicolumn{7}{c}{\textcolor{fine-tuned-color}{\textit{Fine-tuned}}} \\ \hline
\multicolumn{1}{l|}{} &
  \multicolumn{3}{c|}{Seen} &
  \multicolumn{3}{c}{Unseen} \\
\multicolumn{1}{l|}{\multirow{-2}{*}{Method}} &
  Class-Level &
  Intra-Class &
  \multicolumn{1}{c|}{Inter-Class} &
  Class-Level &
  Intra-Class &
  Inter-Class \\ \hline
\rowcolor{fine-tuned-color!\tablealpha} \multicolumn{1}{l|}{TransVG~\cite{deng2021transvg}} &
  0.6294 &
  \underline{0.7213} &
  \multicolumn{1}{c|}{0.6401} &
  0.6984 &
  0.7854 &
  0.7216 \\
\rowcolor{fine-tuned-color!\tablealpha} \multicolumn{1}{l|}{GroundingDino~\cite{liu2023grounding}} &
  \textbf{0.8114} &
  \textbf{0.8011} &
  \multicolumn{1}{c|}{\textbf{0.8077}} &
  \textbf{0.8832} &
  \textbf{0.8554} &
  \textbf{0.8744} \\
\rowcolor{fine-tuned-color!\tablealpha} \multicolumn{1}{l|}{HiVG~\cite{xiao2024hivg}} &
  0.6602 &
  0.731 &
  \multicolumn{1}{c|}{0.7235} &
  0.7373 &
  \underline{0.7892} &
  \underline{0.8176} \\
\rowcolor{fine-tuned-color!\tablealpha} \multicolumn{1}{l|}{Dynamic-MDETR~\cite{shi2024dynamic}} &
  0.7494 &
  0.7166 &
  \multicolumn{1}{c|}{\underline{0.7511}} &
  0.8223 &
  0.7762 &
  \underline{0.8176} \\
\rowcolor{fine-tuned-color!\tablealpha} \multicolumn{1}{l|}{CLIP-VG~\cite{xiao2023clip}} &
  \underline{0.7724} &
  0.6191 &
  \multicolumn{1}{c|}{0.6603} &
  \underline{0.8569} &
  0.6711 &
  0.7433 \\ \hline
\end{tabular}
}
\caption{Visual-grounding performance reported as top-1 bounding-box accuracy at an \textit{IoU} threshold of 0.5.}
\label{tab:grounding}
\vspace{-0.2in}
\end{table*}

\noindent\textbf{Implementation details}. We perform the evaluation only based on the released official models provided by various algorithms on \datasetname and our experiments were conducted using an NVIDIA L20 GPU. For LLAVA~\cite{liu2024visual}, we choose its V1.5-7b version for evaluation. The language model of MiniGPT-4~\cite{zhu2023minigpt} is set to LLaMA-2~\cite{touvron2023llama}. As for the GroundingLMM~\cite{rasheed2023glamm}, we report the results of the models fine-tuned on RefCOCOg dataset~\cite{kazemzadeh2014referitgame} and Visual Genome (VG) dataset~\cite{krishna2017visual}, respectively. For MiniGPT-4 fine-tuning, we train it on 4 NVIDIA A100-40GB for 5 epochs while other training parameters remain the same.

\noindent\textbf{Comparison and analysis}. Based on the results in Table~\ref{tab:instance_capations} and Figure~\ref{fig:quantitative_results}, we summarize the following observations: 1) The generic captioning model predominantly generates coarse phrases that are short and lack domain-specific knowledge. This aligns with the findings in Figure~\ref{fig:caption_length}, where the training caption provided by the general dataset is also brief, in terms of length. The models frequently use everyday vocabulary, such as describing an object as ``a large fish'' instead of the more specific term, ``marlin''. Additionally, the models are prone to misclassifying objects. 2) The models primarily produce image-level captions and struggle to capture fine-grained features, such as morphology, color patterns, and textures. This limitation highlights a significant gap in the ability of general captioning models to support marine-specific tasks effectively. Fine-tuning MiniGPT-4 on \datasetname further demonstrates that domain-specific training enhances image captioning performance. Additional details are provided in the supplementary material.

\subsection{Visual Grounding}
\label{sec:image_grounding}

\noindent\textbf{Experimental settings}. We evaluate visual grounding models under both zero-shot and fine-tuned settings. Specifically, the expert-verified captions from our dataset are used as queries. The algorithms then predict a grounding box, and top-1 accuracy is reported at an Intersection over Union (IoU) threshold of 0.5. We select five models (TransVG~\cite{deng2021transvg}, GroundingDINO~\cite{liu2023grounding}, HiVG~\cite{xiao2024hivg}, Dynamic-MDETR~\cite{shi2024dynamic}, and CLIP-VG~\cite{xiao2023clip}) to assess performance in both zero-shot and fine-tuned scenarios. GroundVLP,~\cite{shen2023groundvlp} (with ALBEF~\cite{li2021alignfusevisionlanguage} employed), a pipeline leveraging pretrained models, is evaluated exclusively in the zero-shot setting.

\noindent\textbf{Implementation details}. To ensure fair comparisons, we adhere strictly to official configuration files and evaluation scripts. For consistency with the GroundingDINO evaluation, which permits only one caption per image, we use the first annotation of each image to construct the testing dataset. In the zero-shot setting, we employ the publicly released models pre-trained on  Flickr30K Entities~\cite{flickrentitiesijcv}, Objects365~\cite{Objects365}, ReferItGame~\cite{kazemzadeh2014referitgame}, based on the setting in the original paper accordingly. 

For the fine-tuned setting, we retrain each model on \datasetname using the same architecture as zero-shot setting and keep other hyperparameters the same.

\noindent\textbf{Comparison and analysis}. From Table~\ref{tab:grounding}, we summarize two observations: 1) Detailed captions facilitate species identification in visual grounding tasks. Unlike object detection, which suffers from significant performance drops in both Intra-Class and Inter-Class settings in Section~\ref{sec:object_detection}, visual grounding tasks demonstrate no notable performance decline, even in the zero-shot setting. This underscores the importance of detailed captions in improving model robustness. 2) Fine-tuning on the \datasetname yields significant performance improvements regarding visual grounding, with top-1 accuracy increasing by at least 10 percentage points for both \textit{seen} and \textit{unseen} categories across all three settings. These results indicate that while detailed captions enable general models to perform reasonably well in unseen marine scenarios, domain-specific supervision provides substantial additional gains.
\section{Discussion and Conclusion}
\noindent\textbf{New benchmark}. \datasetname introduces a comprehensive and diverse benchmark specifically curated for marine research. Designed to advance the evaluation of algorithms for marine visual understanding, it encompasses a broad spectrum of marine species across varied environments, offering a valuable platform for testing and developing new models. 

\noindent\textbf{Limitation}. Despite our efforts to include the most representative marine species, the diversity of marine life far exceeds the current set of categories. We plan to continually expand the dataset to incorporate additional marine objects over time.

\noindent\textbf{Conclusion}. This work presents the first large-scale marine dataset that supports both object recognition and detailed visual understanding. It enables multiple tasks, including \textit{object detection}, \textit{instance captioning}, and \textit{visual grounding}. Our comprehensive evaluation highlights the strengths and limitations of both general-purpose and domain-specific algorithms, providing valuable insights for future research in marine applications.
\section{Acknowledgement}

This project was partially supported by Bridging Horizons: An AI-Powered STEM Learning Initiative in Space and Marine Education under the EdUHK–HKUST Joint Centre for Artificial Intelligence, the HKUST Marine Robotics and Blue Economy Technology Grant, and the Marine Conservation Enhancement Fund (MCEF20107 and MCEF22112).

The authors would also like to express their sincere gratitude to the ``Sustainable Smart Campus as a Living Lab'' (SSC) program at HKUST for its vital support. The program and its dedicated staff not only contributed essential funding and coordination but also fostered the integration of sustainability into campus operations, providing a real-world demonstration of the principles that underpin this research.

{
    \small
    \bibliographystyle{ieeenat_fullname}
    \bibliography{main}
}

\end{document}


\title{ORCA: \underline{O}bject \underline{R}ecognition and \underline{C}omprehension for \\\underline{A}rchiving Marine Species}

\maketitlesupplementary

This supplementary material contains additional details regarding the dataset statistic (Section~\ref{sec:dataset_statistic}), image captioning fine-tuning (Section~\ref{sec:fine_tuning}), and negative caption analysis (Section~\ref{sec:negative_caption_generation}).

\section{Dataset Statistics}
\label{sec:dataset_statistic}

\begin{figure}[h]
    \centering
    \includegraphics[width=\linewidth]{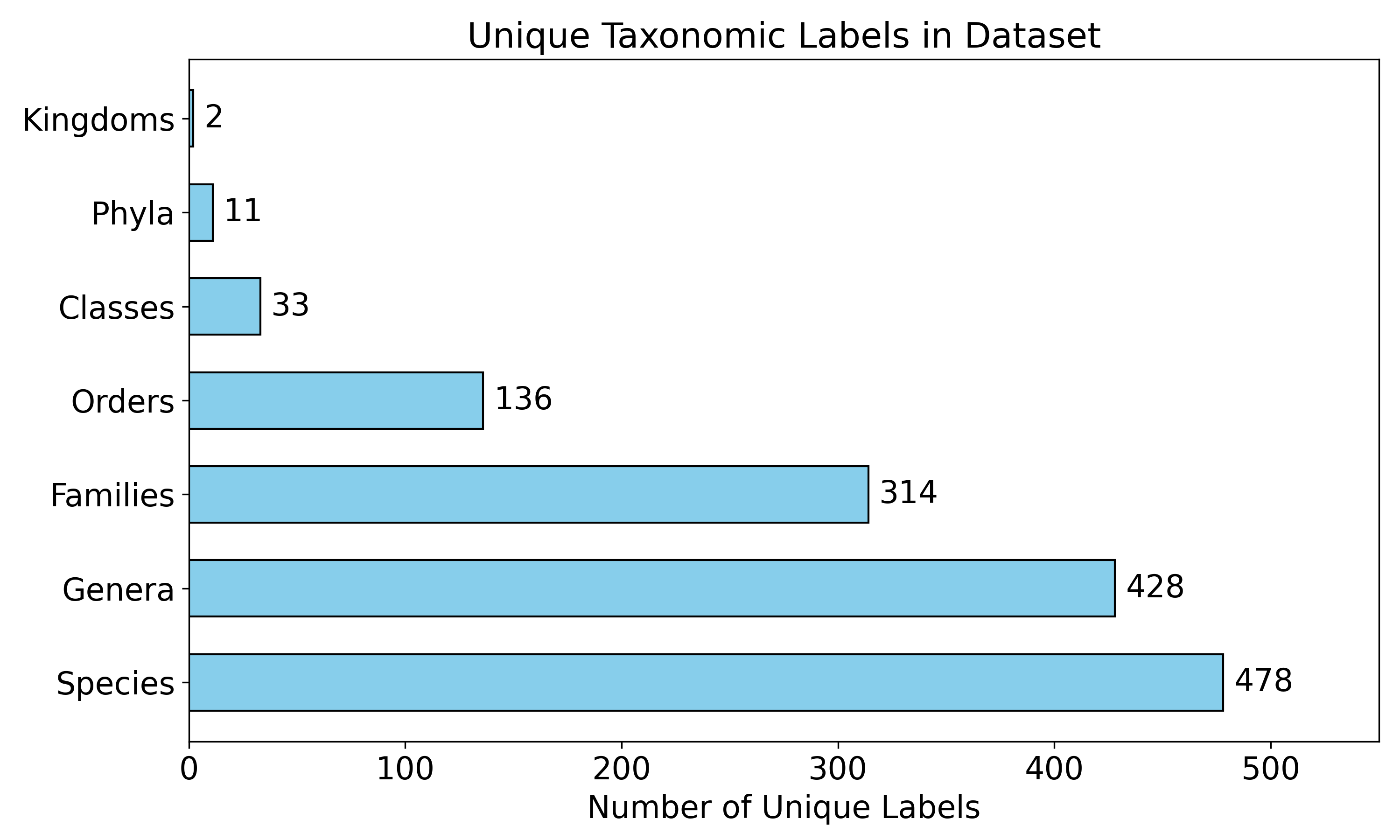}
    \caption{Number of unique labels at each taxonomic level.}
    \label{fig:unique_tax_label}
\end{figure}

\begin{figure}[h]
    \centering
    \includegraphics[width=\linewidth]{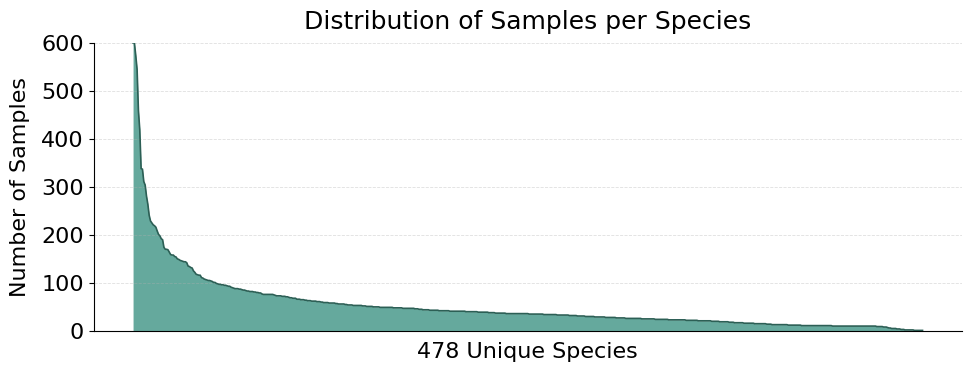}
    \caption{Distribution of sample counts per species in \datasetname{}. The maximum count is clipped to 600 for visualization.}
    \label{fig:species_distribution}
\end{figure}

\begin{figure*}[t]
    \centering
    \includegraphics[width=0.9\linewidth]{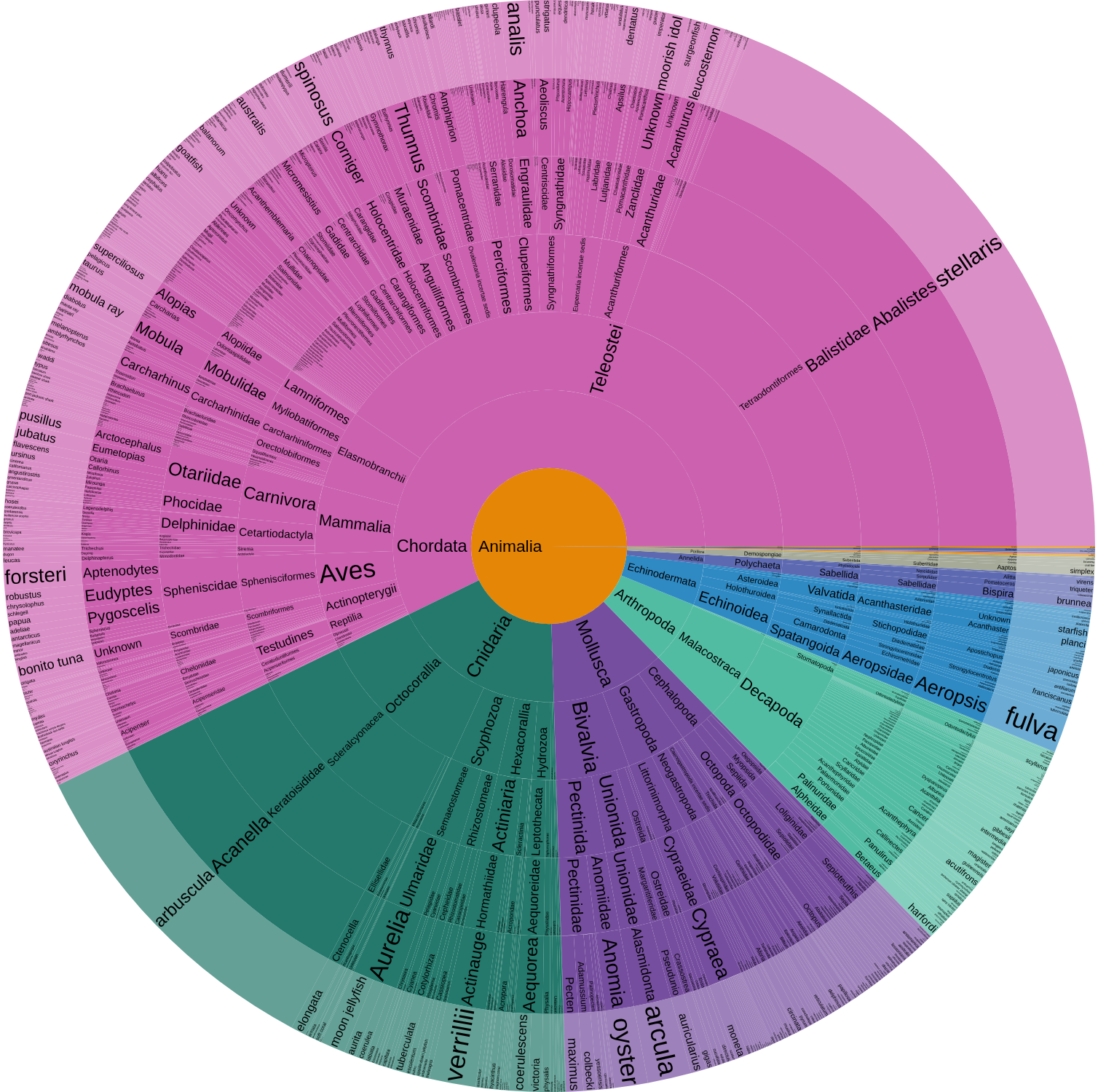}
    \caption{Hierarchical distribution of taxonomic labels in \datasetname{}. Each segment represents a unique label at a given taxonomic rank.}
    \label{fig:tax_label_distribution}
\end{figure*}

This section provides a comprehensive overview of the statistical properties of the \datasetname{} dataset. Figure~\ref{fig:unique_tax_label} summarizes the number of unique taxonomic labels across hierarchical levels, while Figures~\ref{fig:tax_label_distribution} and~\ref{fig:species_distribution} visualize the structural composition and sample distribution within the dataset.

As shown in Figure~\ref{fig:unique_tax_label}, the \datasetname{} dataset captures a wide spectrum of biological diversity, encompassing two kingdoms and 478 distinct species. The hierarchical distribution in Figure~\ref{fig:tax_label_distribution} indicates that most specimens belong to the phylum \textit{Chordata}, with substantial representation from classes such as \textit{Aves} (birds) and \textit{Mammalia} (mammals). Other phyla, including \textit{Mollusca}, \textit{Arthropoda}, and \textit{Cnidaria}, are present to a lesser extent, reflecting the natural sampling bias toward more frequently imaged taxa in ecological and wildlife datasets. This taxonomic heterogeneity underscores both the ecological breadth and inherent imbalance characteristic of large-scale biological image collections.

Figure~\ref{fig:species_distribution} illustrates the distribution of samples across species, revealing a pronounced long-tailed pattern. On average, each species label contains 63.01 images. The dataset exhibits substantial variation in sample counts, from single-instance records of rare species to a maximum of 8,004 samples labeled as “fish.” The aggregated “fish” category reflects practical annotation conventions where specimens of uncertain or undefined identity are grouped under a general label. This natural imbalance poses meaningful challenges for modeling and provides a realistic testbed for evaluating algorithms under imbalanced data scenarios.

\section{Image Captioning Fine-Tuning}
\label{sec:fine_tuning}

We fine-tune MiniGPT-4 on \datasetname to assess the influence of our dataset on image captioning performance. As shown in Table~\ref{tab:minigpt}, fine-tuning on \datasetname yields consistent improvements across nearly all evaluation metrics, demonstrating the dataset’s effectiveness in enhancing the model’s alignment between visual content and textual descriptions. The fine-tuned model achieves higher scores on both semantic and n-gram–based metrics, suggesting that exposure to our dataset enables the generation of captions that are more contextually relevant and descriptively rich.  

It is worth noting that the METEOR score remains unchanged. This can be attributed to our fine-tuning strategy, where only the linear projection layer component of MiniGPT-4 was optimized while the language model remained frozen. Consequently, the semantic vocabulary of the model did not expand beyond the limitations of the original MiniGPT-4 language model, resulting in a stable METEOR score. Nevertheless, fine-tuning with the ORCA dataset allows the model to produce captions that are more accurate and semantically appropriate, aligning more closely with human-authored descriptions. This improvement is reflected in the higher BLEU-4, ROUGE, and CIDEr scores, indicating enhanced ability to convey fine-grained and distinctive visual details.

\begin{table*}
  \centering
  \scalebox{1.0}{
  \begin{tabular}{l|cccccc}
    \toprule
    Method & CLIPScore$\uparrow$ & RefCLIPScore$\uparrow$ & CIDEr$\uparrow$ & BLUE-4$\uparrow$ & METEOR$\uparrow$ & ROUGE$\uparrow$ \\
        \midrule
    Vanilla & 74.48 & 73.43 & 5.72 & 7.18 & 16.90 & 28.03 \\
     Fine-tuned & 77.96$_{\textcolor{blue}{+3.48}}$ & 77.51$_{\textcolor{blue}{+4.08}}$ & 17.36$_{\textcolor{blue}{+11.64}}$ & 14.79$_{\textcolor{blue}{+7.61}}$ & 16.90$_{\textcolor{blue}{+0}}$ & 33.71$_{\textcolor{blue}{+5.68}}$ \\
    \bottomrule 
    \end{tabular}
  }
  \caption{Results of fine-tuning MiniGPT-4 on \datasetname.} 
  \label{tab:minigpt}
\end{table*}
\section{Negative Caption Analysis}
\label{sec:negative_caption_generation}

\begin{table*}[t]
  \centering
  \scalebox{1.0}{
    \begin{tabular}{c|c|l|c}
    \toprule
    & Properties & \multicolumn{1}{c|}{Example} & Number \\
    \midrule
    \multirow{3}{*}{Object} & Classification & This is a yellow \underline{fish}. \textit{vs.} This is a yellow \underline{coral}. & 6,875\\ 
    & Background & The turtle is in the \underline{ocean}. \textit{vs.} The turtle is in the \underline{sky}. & 1,343\\
    & Unexisting & The shark has a long tail. (There is no tail in the image.) & 3,264\\
     \midrule
     \multirow{2}{*}{Relation} & Spatial & This fish is \underline{under} the coral. \textit{vs.} This fish is \underline{on} the coral. & 816\\
     & Action & The penguin is \underline{walking}. \textit{vs.} The penguin is \underline{sitting}. & 938\\
     \midrule
     \multirow{6}{*}{Attribute} 
     & Size & The shark is \underline{large}. \textit{vs.} The shark is \underline{small}. & 271\\
     & Color & This is a \underline{yellow} fish. \textit{vs.} This is a \underline{blue} fish. & 2,031\\
     & Shape & This is a \underline{oval} seashell. \textit{vs.} This is a \underline{triangle} seashell. & 312\\
     & Texture & The seashell is \underline{smooth}. \textit{vs.} The seashell is \underline{rough}. & 321\\
     & Material & The fish is probably made of \underline{plastic}. & 316\\
     & Counting & There are \underline{three} penguins \textit{vs.} There are \underline{four} penguins. & 831\\
    \bottomrule
    \end{tabular}
  }

  \caption{The detailed explanations of the constructed 11 error categories and corresponding data statistics.}
  \label{tab:negative-labels}

\end{table*}

We employ GPT-4 to generate initial image captions, which are subsequently reviewed and refined by marine biologists. Captions identified by the experts as \textit{negative} (incorrect or misleading) are retained to facilitate a detailed examination of captioning errors. A systematic analysis is conducted on these erroneous outputs, categorizing them into 11 distinct types based on their semantic and compositional characteristics. Representative examples and corresponding statistical distributions for each error type are provided in Table~\ref{tab:negative-labels}.

\noindent\textbf{Object-related error}. Our analysis indicates that a substantial proportion of captioning errors stems from object-related issues, particularly those involving fine-grained object classification. This pattern suggests that the model exhibits limited capacity to identify marine species. Additional errors involving nonexistent objects or inaccurate background contexts point to tendencies toward hallucination and contextual misinterpretation, reflecting an overreliance on dataset priors rather than on visual evidence. 

\noindent\textbf{Relation-related error}. Although relation-based errors, which concern spatial or action-level inconsistencies, occur less frequently, they expose deficiencies in the model’s reasoning over inter-object relationships. 

\noindent\textbf{Attribute-related error}. Within the attribute-related error categories, color misclassification emerges as the most prevalent, underscoring the model’s sensitivity to variations in illumination, shading, and surface textures. Collectively, these findings highlight the need for improved visual grounding, compositional reasoning, and quantitative perceptual mechanisms to enhance both the semantic precision and contextual reliability of image captioning systems.